\documentclass[10pt,twocolumn,letterpaper]{article}

\usepackage{iccv}
\usepackage{times}
\usepackage{epsfig}
\usepackage{graphicx}
\usepackage{amsmath}
\usepackage{amssymb}
\usepackage{kotex}
\usepackage[flushleft]{threeparttable}
\usepackage[linesnumbered,ruled,vlined]{algorithm2e}
\usepackage{multirow}
\usepackage{makecell}
\usepackage[symbol]{footmisc}
\usepackage{float}
\usepackage{adjustbox}
\usepackage[T1]{fontenc}
\usepackage{etoolbox}
\usepackage{caption}
\usepackage{flushend}
\usepackage{adjustbox}
\usepackage{threeparttable}
\usepackage{multirow}
\usepackage{makecell}
\usepackage[pagebackref=true,breaklinks=true,letterpaper=true,colorlinks,bookmarks=false]{hyperref}
\usepackage{footnote}
\iccvfinalcopy 

\def\iccvPaperID{2336} 

\ificcvfinal\fi
\begin{document}

\title{Frame-to-Frame Aggregation of Active Regions in Web Videos for Weakly Supervised Semantic Segmentation}

\author{Jungbeom Lee$^1$ ~~~~~~~ Eunji Kim$^1$ ~~~~~~~  Sungmin Lee$^1$  ~~~~~~~   Jangho Lee$^1$ ~~~~~~~  Sungroh Yoon$^{1, 2, }$\thanks{Correspondence to: Sungroh Yoon <sryoon@snu.ac.kr>.}\\
$^1$ Department of Electrical and Computer Engineering, Seoul National University, Seoul, South Korea\\
$^2$ ASRI, INMC, ISRC, and Institute of Engineering Research, Seoul National University\\
{\tt\small \{jbeom.lee93, kce407, simonlee0810, ubuntu, sryoon\}@snu.ac.kr}}

\maketitle

\begin{abstract}

When a deep neural network is trained on data with only image-level labeling, the regions activated in each image tend to identify only a small region of the target object. 
We propose a method of using videos automatically harvested from the web to identify a larger region of the target object by using temporal information, which is not present in the static image. The temporal variations in a video allow different regions of the target object to be activated.
We obtain an activated region in each frame of a video, and then aggregate the regions from successive frames into a single image, using a warping technique based on optical flow. The resulting localization maps cover more of the target object, and can then be used as proxy ground-truth to train a segmentation network. This simple approach outperforms existing methods under the same level of supervision, and even approaches relying on extra annotations. Based on VGG-16 and ResNet 101 backbones, our method achieves the mIoU of 65.0 and 67.4, respectively, on PASCAL VOC 2012 test images, which represents a new state-of-the-art.

\end{abstract}

\begin{figure}[t]
  \centering
  \includegraphics[width=1.0\linewidth]{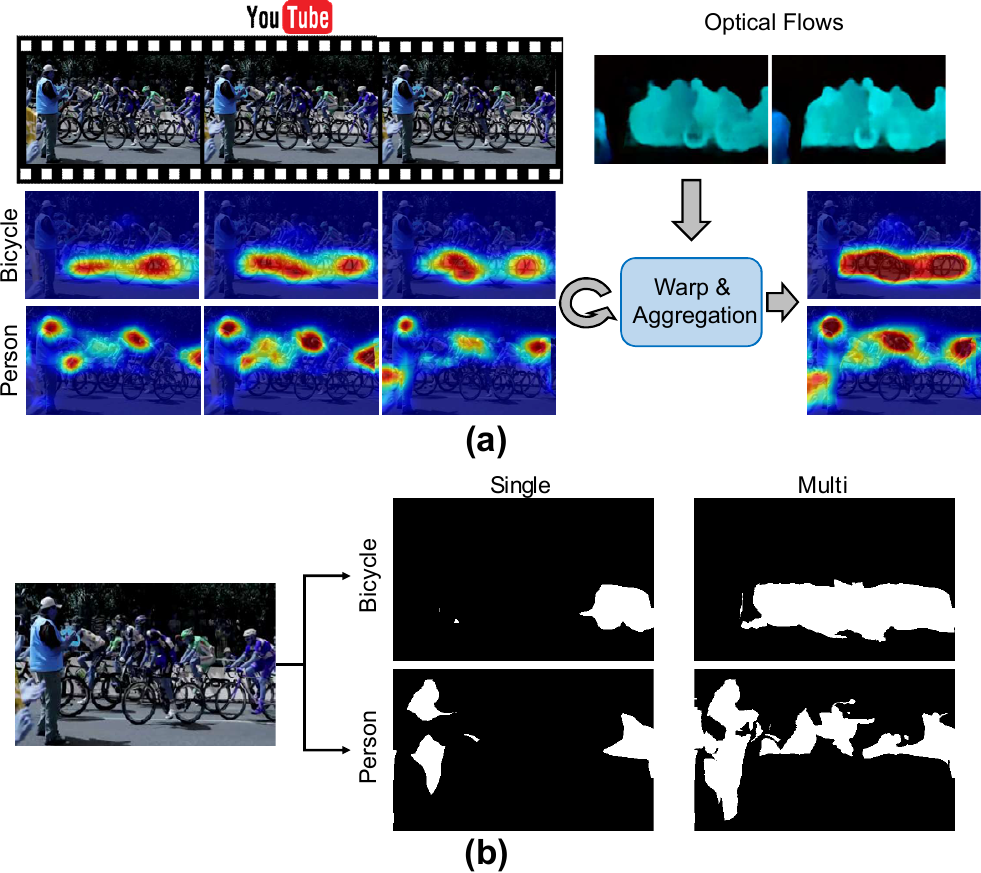} \\[-0.7em]
  \caption{\label{fig_1} (a) Our method discovers activated regions from each frame and aggregates them into a single frame using a warping technique based on optical flow. (b) Comparison of generated proxy ground truth masks obtained from a single frame (1st row) and aggregated for multiple frames (2nd row). The latter covers larger region of the target objects because it embodies information from several frames. \vspace{-10pt}}
\end{figure}
\vspace{-10pt}
\section{Introduction}

Semantic segmentation is one of the most important tasks in computer vision and one that has made tremendous progress with fully annotated pixel-level labels~\cite{zhao2017pyramid,chen2014semantic}.
However, real applications of semantic image segmentation require a large variety of object classes and a great deal of labeled data for each class, and labeling pixel-level annotations is laborious. 
This problem can be addressed by weakly supervised methods that use more easily obtainable annotations such as scribbles, bounding boxes, or image-level tags.

Weakly supervised semantic segmentation methods have evolved rapidly. Scribble supervision~\cite{tang2013discriminative} can now achieve 97\% of the performance of manually supervised semantic segmentation with the same backbone network, and even the weakest image-level supervision~\cite{lee2019ficklenet} can produce 90\% of that performance.
However, the rate at which segmentation techniques relying on weak annotations are improving is declining rapidly.
For instance, under image-level supervision, a 7.9\% improvement on PASCAL VOC 2012 validation images~\cite{everingham2010pascal} was achieved between 2016 and 2017~\cite{kolesnikov2016seed, chaudhry2017discovering}, 1.8\% between 2017 and 2018~\cite{wei2018revisiting}, but only 0.8\% since then~\cite{lee2019ficklenet}.

Most methods of weakly supervised segmentation use localization maps activated by a classifier (e.g. a CAM~\cite{zhou2016learning}) as a proxy ground truth to train their segmentation network. However, because such localization maps are coarse and therefore activate only part of a target object, they provide a poor version of proxy ground truth. 
Some researchers have tried to address this problem by erasing discriminative part of the target object~\cite{wei2017object, hou2018self} or introducing different sizes of receptive fields~\cite{wei2018revisiting, lee2019ficklenet, lee2018robust}, but these developments encounter limitations because of the finite information that can be obtained from the image-level annotations, resulting in a coarse proxy ground truth.
Of course, stronger methods of supervision, such as instance-level saliency~\cite{hu2018associating}, scribbles~\cite{tang2018normalized, tang2018regularized}, or bounding boxes~\cite{dai2015boxsup} can be used, but the extra supervision required makes it harder to expand object classes or move to a new domain.

One way of reducing the effort required for strong supervision is to use so-called webly supervised segmentation, which makes use of the vast amount of image and video data on the web, which can be used in combination with other existing weakly annotated data.
However, the data obtained from the web may be of low quality or may not depict any objects corresponding to the search term.
Attempts have been made to improve the quality of such data by filtering method using knowledge learned from images with accurate image-level annotations (e.g. the PASCAL VOC 2012 dataset~\cite{everingham2010pascal}), or by using existing image segmentation techniques such as GrabCut~\cite{rother2004grabcut}.
Despite labeling issues, some of the sophisticated webly supervised methods have shown remarkable performance~\cite{shen2018bootstrapping, hong2017weakly, hou2018webseg, jin2017webly}.
However, additional usage of web images brings the same level of coarseness as existing weakly annotated data, so the performance improvements using web images are merely due to the effects of data augmentation.
On the other hand, the temporal variations in video allow a classifier to activate different regions of the target object, so that video offers the possibility of obtaining better pixel-level annotations than static images.


We propose a method of collecting regions activated in different frames, using a warping technique based on optical flow.
Optical flow provides pixel-level displacements between two successive frames, making it possible to deduce which parts of the frames correspond.
Warping the activated areas in the first frame into the second frame determines which areas of the second frame should be activated. These warped areas can be aggregated with the parts of the second frame which are activated on their own merits.
We repeat this step so that the areas activated in several frames become available in a single frame.
Figure~\ref{fig_1}(a) shows how this works on an example, and Figure~\ref{fig_1}(b) demonstrates the resulting increase in the area of the activated regions, and how this corresponds more clearly to the ground truth of the original image.
In addition, the availability of multiple frames in a video allows more effective filtering to refine inaccurate labels (see Section~\ref{filtering}).

Existing methods of webly supervised segmentation depend on off-the-shelf segmentation techniques~\cite{hong2017weakly, shen2018bootstrapping} such as GrabCut~\cite{rother2004grabcut}, complicated optimization of energy function~\cite{hong2017weakly, tokmakov2016weakly}, or heuristic constraints~\cite{shen2018bootstrapping} to generate a proxy ground truth of the data obtained from the web. None of these are needed by our technique.

The main contributions of this paper can be summarized as follows:
\begin{itemize}
	\item[$\bullet$] We propose simple data filtering and incremental warping techniques which allow web videos to be used as an additional data source in weakly supervised semantic image segmentation.
	
	\item[$\bullet$] We empirically demonstrate that our method of processing web video data improves the performance of several methods of weakly supervised segmentation.
	
	\item[$\bullet$] Our technique significantly outperforms other state-of-the-art methods on the Pascal VOC 2012 benchmark in both weakly supervised and webly supervised settings.

\end{itemize}

\section{Related Work}
Over the past four years, improvements of over 20$\%$ in the PASCAL VOC 2012 benchmark have been achieved by fully supervised semantic segmentation~\cite{long2015fully, yu2018learning}.
However, because of the difficulty of obtaining pixel-level annotations of all the types of image and classes of object encountered in real applications, it is difficult to apply semantic segmentation widely. 
Weakly supervised semantic segmentation methods have been proposed to address this problem, and they have achieved promising performance (Section~\ref{weaklysuprelated}). Webly supervised semantic segmentation methods using images or videos obtained by web crawlers have been introduced as a way of closing the gap between the performance of weakly and fully supervised methods (Section~\ref{weblysuprelated}).

\subsection{Weakly supervised semantic segmentation}\label{weaklysuprelated}
The goal of weakly supervised semantic segmentation is to train a image segmentation network with relatively imprecise delineations of the objects to be recognized. Weak supervision can take the form of scribbles~\cite{tang2018regularized, tang2018normalized}, bounding boxes~\cite{dai2015boxsup}, or image-level tags, which is the approach on which will now focus.

Most methods of image-level annotation are based on a class activation map (CAM)~\cite{zhou2016learning}. However, it is widely known that a CAM only identifies small discriminative parts of a target object~\cite{zhang2018adversarial, huang2018weakly, wei2017object}. Thus, the localization maps obtained by CAM are insufficiently complete to be used as a proxy ground truth to train a segmentation network.
Several techniques have been proposed to expand these activated regions to the whole target object.
Erasing methods~\cite{kim2017two, wei2017object, hou2018self} prevent a classifier from focusing solely on the discriminative parts of objects by removing those regions.
Other methods construct CAMs that embody the multi-scale context of the target object, by means of different sizes of receptive fields.
MDC~\cite{wei2018revisiting} computes CAMs from features which have different receptive fields, realized by several convolutional blocks dilated at different rates.
Pyramid Grad-CAM~\cite{lee2018robust} collects features from each of several densely connected layers and merges the resulting localization maps. FickleNet~\cite{lee2019ficklenet} selects features stochastically by using modified dropout technique: this not only prevents the classifier from concentrating solely on the discriminative part, but also uses a different receptive field for every inference.

Region growing methods try to expand regions of the target object, starting from the initial CAM as a seed.
AffinityNet~\cite{ahn2018learning} and CIAN~\cite{fan2018cian} consider pixel-level semantic affinities, which identify relationship of pixels, and grow regions based on those affinities. SEC~\cite{kolesnikov2016seed} and DSRG~\cite{huang2018weakly} progressively refine initial localization maps during the training of their segmentation network by means of resulting segmentation maps during training time, which are refined using a conditional random field (CRF).

\subsection{Webly supervised semantic segmentation}\label{weblysuprelated}
Along with the growth of weakly supervised semantic segmentation, some researchers have attempted to improve the performance of weakly supervised methods using additional images or videos obtained from the web. WebS-i2~\cite{jin2017webly} collects two types of web data: images showing objects of the target class against a white backgrounds and images containing common backgrounds without any objects of interest class. It then trains the segmentation network by means of an iterative refinement process on realistic images whose weak annotations are accurately labelled.
Web-Crawl~\cite{hong2017weakly} selects videos from YouTube by examining thumbnails, and segments them by spatio-temporal graph-based optimization.
Bootstrap-Web~\cite{shen2018bootstrapping} finds images which are expected to be easy to segment using heuristic constraints such as the sizes of objects, and exchanges knowledge between two networks; one is trained by filtered easy-to-segment web images and the other is trained by realistic hard-to-segment images.
All these methods improve segmentation performance to some extent, but they are largely dependent on complicated optimization methods, on heuristic constraints, or on off-the-shelf segmentation methods such as GrabCut~\cite{rother2004grabcut}






\section{Proposed Method}

Our goal is to train a segmentation network with weakly annotated image data $\mathcal{I}$ and web video data $\mathcal{V}$. While the image-level labels of each image in $\mathcal{I}$ have been manually annotated, $\mathcal{V}$ is annotated with noisy video-level tags because it has been collected from the web by searching with the name of each object class as the search term. Our training procedure has the following steps: A deep neural network is trained on $\mathcal{I}$ to identify classes of object (Section~\ref{classification}). 
After it has been trained, the network processes the videos in $\mathcal{V}$, and the classification results can be used to filter out the irrelevant frames or update the labels of the videos $\mathcal{V}$ (Section~\ref{filtering}).
The proxy ground truths for selected sequences of frames are then generated using an incremental warping method (Section~\ref{proxygt}). Finally, the proxy ground truth is used to train a segmentation network (Section~\ref{segnet}). The overall procedure is shown as Algorithm~\ref{Procedure}.

\begin{algorithm}
\footnotesize
\KwIn{Image dataset $\mathcal{I}$, web video dataset $\mathcal{V}$}
Train a classifier using $\mathcal{I}$ \hspace*{\fill} Sec.~\ref{classification}\\
$\mathcal{\hat{V}} \gets$ Data filtering for $\mathcal{V}$ \hspace*{\fill} Sec.~\ref{filtering}\\
\textbf{Generating Proxy Ground Truth:}\hspace*{\fill} Sec.~\ref{proxygt}\\
\qquad $\mathcal{M}_{\mathcal{I}}, \mathcal{M}_{\mathcal{\hat{V}}} \gets$ Masks from CAMs on $\mathcal{I}$ and $\mathcal{\hat{V}}$\hspace*{\fill} Sec.~\ref{infercam}\\
\qquad $\hat{\mathcal{M}}_{\mathcal{\hat{V}}} \gets$ Union by incremental warping $\mathcal{M}_{\mathcal{\hat{V}}}$\hspace*{\fill} Sec.~\ref{warping}\\

\textbf{Training Segmentation Network:}\hspace*{\fill} Sec.~\ref{segnet}\\
\qquad $L_{\mathcal{I}} \gets$ Compute segmentation loss using $(\mathcal{I}, \mathcal{M}_{\mathcal{I}}$)\\
\qquad $L_{\mathcal{\hat{V}}} \gets$ Compute segmentation loss using $(\mathcal{\hat{V}}, \hat{\mathcal{M}}_{\mathcal{\hat{V}}}$)\\
\qquad Update segmentation network by $L_{\mathcal{I}} + L_{\mathcal{\hat{V}}}$
\caption{Overall Procedure}\label{Procedure}
\end{algorithm}
\vspace{-8pt}
\subsection{Learning with precise labels}\label{classification}
We train a deep convolutional neural network using $\mathcal{I}$, which has precise image-level multi-class labels. In order to obtain class activation maps (CAM) (Section~\ref{proxygt}), we modify the VGG-16 network~\cite{simonyan2014very} to be fully convolutional, by removing all the fully connected layers and adding additional convolutional layers so that the number of channels of the final output feature corresponds to the number of classes of interest. We then apply global average pooling (GAP) and a sigmoid function to the feature output by the network, so as to obtain a score for each class. We use these results to update the parameters of the classifier by means of a sigmoid cross-entropy loss function, of a sort widely used for multi-label classification.

\subsection{Data filtering for noisy dataset}\label{filtering}
Images or videos which are expected to depict a certain class of object can be obtained from the web by searching, using the name of that class as the search term.
That name is then used to label the images or videos that are acquired.
But not all the resulting images or videos will actually show an object of that class, and many of them will contain objects of classes other than that corresponding to the search term. For example, a video obtained by searching for "horse" may show a person riding a horse; but it will just be labeled "horse", which is the search term.

Most webly supervised segmentation methods use the knowledge obtained from precisely annotated image data to eliminate web images or videos which do not depicit any objects corresponding to the search term. 
For example, Bootstrap-Web~\cite{shen2018bootstrapping} uses SEC~\cite{kolesnikov2016seed}, which is trained by precisely annotated data, to obtain pixel-level class masks from web images. It discards images which have too few or too many pixels assigned to an object of the search-term class in the corresponding mask.
Web-Crawl~\cite{hong2017weakly} deals specifically with videos, and rejects a video if it has fewer than 5 frames with classification scores for its search term that achieve a threshold.
These methods work on the assumption that there is only a single class of object in an image or video. But automatically collected images or videos can be expected to contain objects of many classes. In PASCAL VOC 2012 data, more than 36\% of training images are annotated with more than one class.

We therefore introduce an incremental multi-class filtering and label refining method which considers several successive frames of a video. 
We can eliminate videos which show no objects of interest and correct inaccurate labels much more effectively than these processes can be performed on single images.
When the same class of object is found in consecutive frames of a video, it becomes increasing likely that an object of that class is actually depicted. This means that labels can be assigned with more confidence to a video than to an image, and multi-class labelling becomes more feasible.

A classifier trained in Section~\ref{classification} processes the videos in $\mathcal{V}$, and infers the object classes present in each frame, which are taken to be those with a score larger than a threshold $\tau$. The set $\mathcal{C}$ of labels of these classes is then attached to each frame.
We add a sequence of $K$ frames to the filtered video set $\hat{\mathcal{V}}$ if those $K$ frames all show objects from the same set of classes $\mathcal{C}$, and one of those classes corresponds to the search term. We eliminate frames that do not satisfy the above conditions.

\begin{figure}[t]
  \centering
  \includegraphics[width=1.0\linewidth]{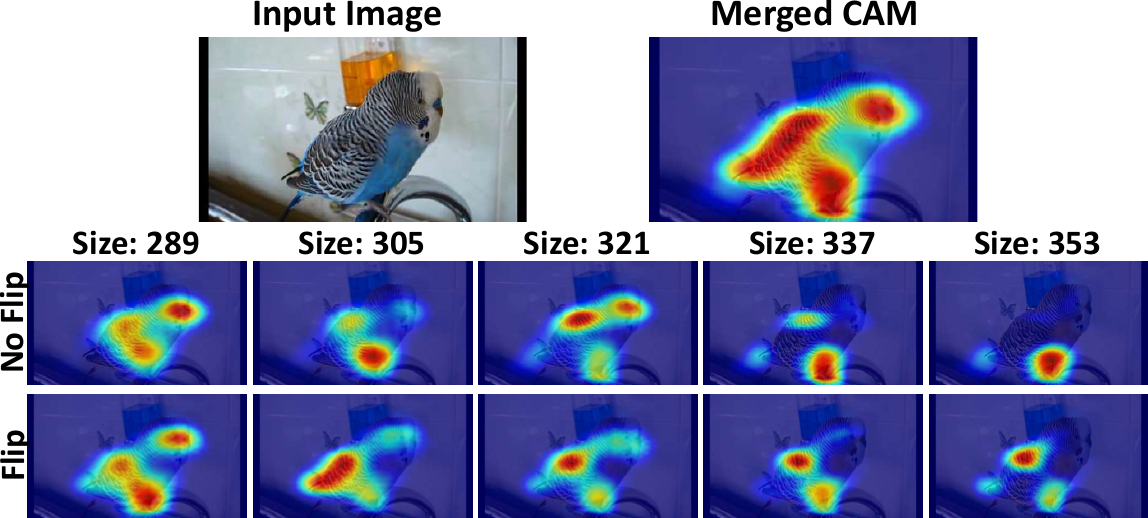} \\[-0.7em]
  \caption{\label{fig_camsample} CAMs obtained by flipping and rescaling an input image. Rescaling changes the granularity at which the content of the image is considered. Different regions tend to be activated by flipped image.\\[-2em]}
\end{figure}

\subsection{Generating proxy ground truth}\label{proxygt}
We now have a set of images $\mathcal{I}$, with accurate image-level labels, and a set of web videos $\hat{\mathcal{V}}$ with labels which are more accurate than those originally obtained from the search terms. 
We now describe the creation of a proxy ground truth for the frames in $\hat{\mathcal{V}}$, and its use to train a segmentation network.

\subsubsection{Inference localization maps}\label{infercam}
\vspace{-0.3em}
We use a CAM~\cite{zhou2016learning} to obtain a localization map for each class of object in the images. 
Zhang \textit{et al.}~\cite{zhang2018adversarial} showed empirically, and also proved mathematically that the $c-$th channel of the last feature becomes the CAM for the $c-$th class, within a fully convolutional network of the type described in Section~\ref{classification}.
\begin{figure}[t]
  \centering
  \includegraphics[width=1.0\linewidth]{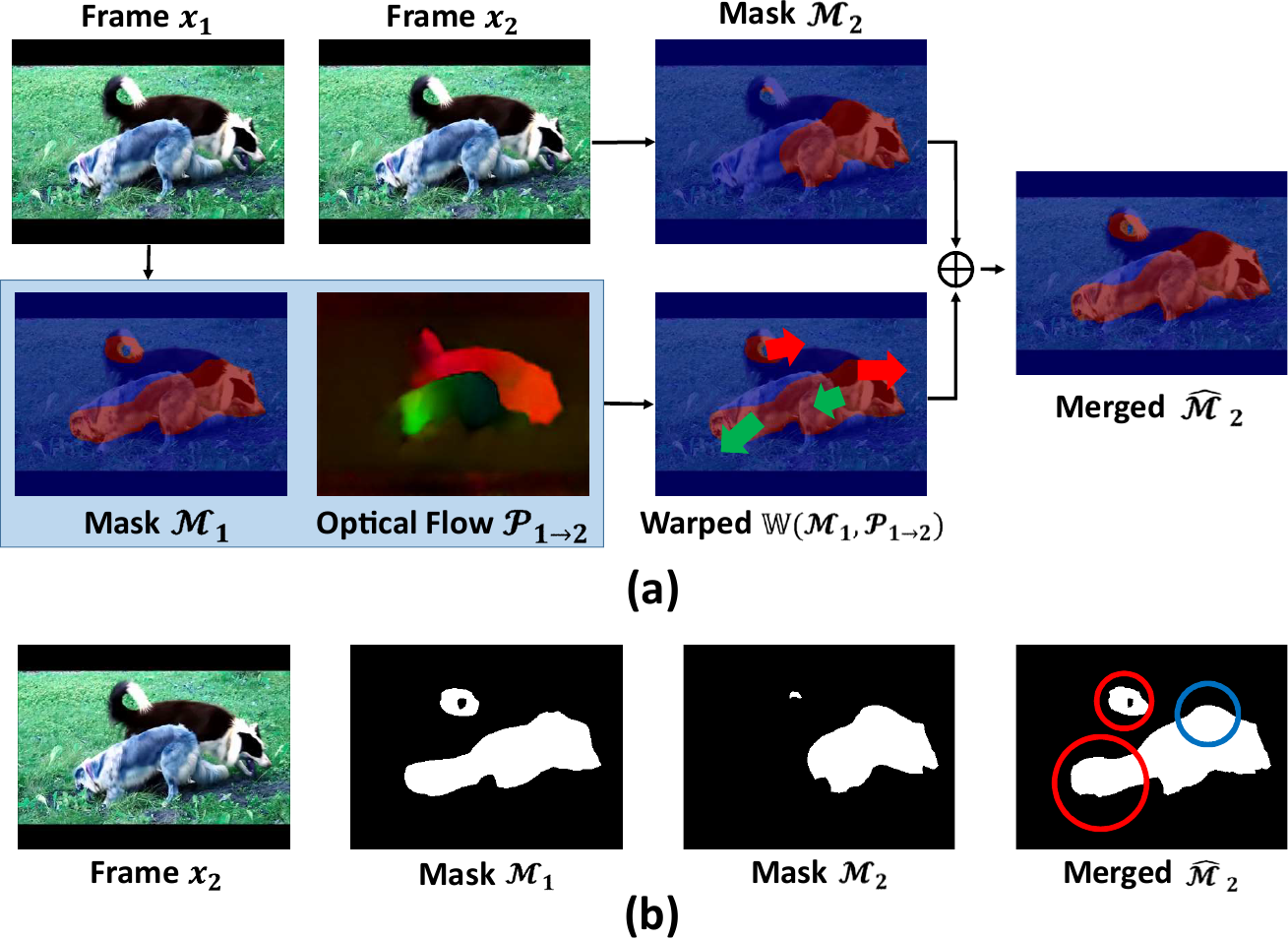} \\[-0.7em]
  \caption{(a) Conceptual description of single-step aggregation. The mask of activated regions $\mathcal{M}_1$ is warped to $\mathbb{W}(\mathcal{M}_1, \mathcal{P}_{1\rightarrow 2})$, and aggregated with the mask $\mathcal{M}_2$, to generate $\hat{\mathcal{M}}_2$. (b) Examples of generated masks. The red circles denote the regions masked only by $\mathcal{M}_1$, and the blue circle denotes the region masked only by $\mathcal{M}_2$. The mask $\hat{\mathcal{M}}_2$ covers more of the target objects than $\mathcal{M}_1$ and $\mathcal{M}_2$.}\vspace{-1em}
  \label{warp_samples}
\end{figure}
We obtain CAMs from images which are scaled or flipped horizontally.
After applying inverse transformations to the CAMs, we select the maximum value at each pixel over all the maps. Figure~\ref{fig_camsample} shows some examples of the effect of flipping and scaling on CAMs. 
The CAM from a small image provides a coarse localization, and a CAM from a larger image identifies more detail of the object.
Additionally, different regions are activated by flipped image. \\[-1.8em]

\vspace{-0.3em}

\subsubsection{Incremental warping of localization maps}\label{warping}
\vspace{-0.3em}
It is widely known that a CAM only identifies the small discriminative part of a target object~\cite{zhang2018adversarial, huang2018weakly, wei2017object}.
Our method obtains information about larger regions of a target object by merging information from successive frames of a video. 
The masks that indicate different regions of the target object are obtained from successive frames, and then aggregated into a single mask by considering optical flow.
An analysis of optical flow between two successive frames provides a warp, which encodes the displacements which relate the pixels in one frame to those in the other.
By warping the mask obtained from a frame to the following frame, we can transfer the activated region of the first frame to the second frame, allowing the union of the activated regions of both frames to be considered in a single image.

Let $\mathcal{X} \in \hat{\mathcal{V}}$ be a video containing $K$ frames $\{x_i\}_{i=1}^K$. 
We compute masks $\{\mathcal{M}_i^c\}_{i=1}^K$ for each class $c$ in $\mathcal{C}$ by thresholding the localization maps of the $\{x_i\}_{i=1}^K$ with the threshold $\theta_{f}$. We obtain optical flows $\{\mathcal{P}_{i\rightarrow i+1}\}_{i=1}^{K-1}$ between each pair of successive frames.

We will now consider a single aggregation step, which combines the masks $\mathcal{M}_1^c$ and $\mathcal{M}_2^c$ obtained from successive frames $x_1$ and $x_2$, for a single class $c$. 
Let $\mathbb{W}(I, f)$ be the function which warps an image $I$ following the flow field $f$, using bilinear interpolation. 
If $\mathcal{P}_{1\rightarrow 2}$ is the flow field between $x_1$ and $x_2$, then we can use this function to warp $\mathcal{M}_1^c$ to the space of $x_2$.
The warped mask $\mathbb{W}(\mathcal{M}_1^c, \mathcal{P}_{1\rightarrow 2})$ expresses regions of $x_2$ which correspond to the activated regions of $x_1$. 
An aggregated mask $\hat{\mathcal{M}}_2^c$ can then be obtained as the union of the warped mask from $x_1$ and the mask $\mathcal{M}_2^c$ obtained from $x_2$.
\begin{equation}\label{warp}
   \hat{\mathcal{M}_2^c} = \mathcal{M}_2^c \cup \mathbb{W}(\mathcal{M}_1^c, \mathcal{P}_{1 \rightarrow 2}).
\end{equation}
We repeat this procedure for the remaining annotated object classes in $\mathcal{C}$, and call the result $\hat{\mathcal{M}}_2$. 
This procedure is illustrated in Figure~\ref{warp_samples}(a), and Figure~\ref{warp_samples}(b) shows that a unioned mask map $\hat{\mathcal{M}}_2$ contains the activated regions from both frame $x_1$ and $x_2$.
$\hat{\mathcal{M}}_2$ can then be warped using the optical flow $\mathcal{P}_{2 \rightarrow 3}$ and aggregated with $\mathcal{M}_3$, producing $\hat{\mathcal{M}}_3$.
By repeating this procedure until the $K-$th frame, we can obtain $\hat{\mathcal{M}}_K$, which contains the union of all the activated regions from all $K$ frames. Aggregated masks from all the videos in $\hat{\mathcal{V}}$ can then be used as a proxy ground truth to train a segmentation network.

\subsection{Segmentation Network}\label{segnet}
Many weakly supervised segmentation methods are built upon existing weakly supervised segmentation networks. For example, MDC~\cite{wei2018revisiting} is based on a slightly modified version of the AE-PSL~\cite{wei2017object}, and GAIN~\cite{li2018tell}, TPL~\cite{kim2017two}, and Boost-Web~\cite{shen2018bootstrapping} are based on SEC~\cite{kolesnikov2016seed}. 
We used FickleNet~\cite{lee2019ficklenet}, but we also experimented with two other popular weakly supervised semantic segmentation networks: SEC~\cite{kolesnikov2016seed} and DSRG~\cite{huang2018weakly}. 
All these segmentation networks are trained using the proxy ground truth of $\hat{\mathcal{V}}$ generated by the method described in Section~\ref{proxygt}. 
The background of $\hat{\mathcal{V}}$ is identified by saliency detection ~\cite{hou2017deeply} and includes all pixels with saliency values lower than $\theta_b$.
When each segmentation network is trained, we use the proxy ground truth of $\mathcal{I}$ provided by the authors of each segmentation method.

For each iteration in training time, we create a batch; half of the elements in the batch come from $\mathcal{I}$, and the other half from $\hat{\mathcal{V}}$.
We obtain the segmentation losses $L_{\mathcal{I}}$ and $L_{\mathcal{\hat{V}}}$ from the data from $\mathcal{I}$ and $\hat{\mathcal{V}}$ respectively.
We then update the segmentation network by $L_{\mathcal{I}} + L_{\mathcal{\hat{V}}}$.
Since $\mathcal{I}$ and $\hat{\mathcal{V}}$ may have different data distributions, we perform the domain adaptation according to~\cite{hong2017weakly}: the segmentation network is trained using both $\mathcal{I}$ and $\hat{\mathcal{V}}$ to predict segmentation masks for the images from $\mathcal{I}$, and those maps are used as a proxy ground truth to fine-tune the network.

\section{Experiments}

\subsection{Experimental Setup}\label{impledetail}
\noindent\textbf{Image Dataset:} We conducted experiments on the PASCAL VOC 2012 image segmentation benchmark~\cite{everingham2010pascal}, which contains 20 foreground object classes and one background class. Using the same protocol as other work on weakly supervised semantic segmentation, we trained our network using augmented 10,582 training images with image-level annotations. We determined mean intersection-over-union (mIoU) values for 1,449 validation images and 1,456 test images. The results for the test images were obtained from the official PASCAL VOC evaluation server. 

\noindent\textbf{Web Video Dataset:}
Starting with the Web-Crawl~\cite{hong2017weakly} dataset, we filtered out irrelevant frames and refine inaccurate labels with a threshold $\tau=0.9$, and aggregated masks from $K=5$ frames into a single mask, yielding 15,000 final samples for training a segmentation network. If several sets of frames can be selected from each video in Section~\ref{filtering}, we chose only one to avoid similarity of samples.
We obtained optical flows using PWC-Net~\cite{sun2018pwc}. 
The foreground and background thresholds, $\theta_f$ and $\theta_b$, were set to 0.2 and 0.12 respectively.

\noindent\textbf{Classification Network:}
Our classifier is based on the VGG-16 network~\cite{simonyan2014very}, pre-trained using the Imagenet~\cite{deng2009imagenet} dataset. The VGG-16 network was modified by removing all the fully connected layers and the last pooling layer, and we replaced the convolutional layers of the last block with convolutions dilated with a rate of 2. We added two convolutional layers with 1024 channels and a kernel size of 3 with 2D dropout~\cite{srivastava2014dropout}.

\noindent\textbf{Segmentation Network:}
As already stated, we experimented with FickleNet~\cite{lee2019ficklenet}, SEC~\cite{kolesnikov2016seed}, and DSRG~\cite{huang2018weakly}.
We followed the settings recommended by the authors of those methods, except for the optional domain adaptation process, during which the learning rate was reduced to 0.01 from the default learning rate.

\noindent\textbf{Reproducibility:} PyTorch~\cite{paszke2017automatic} was used for training the classifier, extracting CAMs, and obtaining optical flow~\cite{sun2018pwc}, and we used the Caffe deep learning framework~\cite{jia2014caffe} in the segmentation step.

\begin{table}[htbp]
  \centering
  \caption{Comparison of the performance of weakly supervised segmentation
methods using VGG16-based segmentation model on VOC 2012 validation and test sets.\\[-2em]}
    \begin{tabular}{lccc}
    \Xhline{1pt}\\[-0.95em]
    Methods  & ~~\textit{val}~~ & \textit{test} \\
    \hline\hline 
    \\[-0.9em]
    
    \multicolumn{3}{l}{Supervision: Image-level and additional annotations} \\
    $\text{MIL-seg}_{\text{~~CVPR '15}}$~\cite{pinheiro2015image}   & 42.0    & 40.6 \\
    $\text{STC}_{\text{~~TPAMI '17}}$~\cite{wei2017stc}      & 49.8  & 51.2 \\
    $\text{TransferNet}_{\text{~~CVPR '16}}$~\cite{hong2016learning} ~~~~~~~~&  52.1  & 51.2 \\
      $\text{AISI}_{\text{~~ECCV '18}}$~\cite{hu2018associating} & 61.3  & 62.1 
      \\\\[-0.9em]
    
    \hline
    \\[-0.9em]
    \multicolumn{3}{l}{Supervision: Image-level annotations only}\\
    $\text{SEC}_{\text{~~ECCV '16}}$~\cite{kolesnikov2016seed}    & 50.7  & 51.1 \\
    $\text{CBTS-cues}_{\text{~~CVPR '17}}$~\cite{roy2017combining}    & 52.8  & 53.7 \\
    $\text{TPL}_{\text{~~ICCV '17}}$~\cite{kim2017two}   & 53.1  & 53.8 \\
    $\text{AE\_PSL}_{\text{~~CVPR '17}}$~\cite{wei2017object}    & 55.0    & 55.7 \\
    $\text{DCSP}_{\text{~~BMVC '17}}$~\cite{chaudhry2017discovering}  & 58.6 & 59.2\\
    $\text{MEFF}_{\text{~~CVPR '18}}$~\cite{ge2018multi}  & - & 55.6\\
    $\text{GAIN}_{\text{~~CVPR '18}}$~\cite{li2018tell}    & 55.3  & 56.8 \\
    $\text{MCOF}_{\text{~~CVPR '18}}$~\cite{wang2018weakly}    & 56.2  & 57.6 \\
    $\text{AffinityNet}_{\text{~~CVPR '18}}$~\cite{ahn2018learning}    & 58.4  & 60.5 \\
    $\text{DSRG}_{\text{~~CVPR '18}}$~\cite{huang2018weakly}   & 59.0    & 60.4 \\
    $\text{MDC}_{\text{~~CVPR '18}}$~\cite{wei2018revisiting}  & 60.4  & 60.8 \\
    $\text{SeeNet}_{\text{~~NIPS '18}}$~\cite{hou2018self}   & 61.1 & 60.7 \\
    $\text{FickleNet}_{\text{~~CVPR '19}}$~\cite{lee2019ficklenet}   & 61.2 & 61.9 \\
    Ours    &  \textbf{63.9}  &  ~~\textbf{65.0}\\
    \Xhline{1pt}
    \end{tabular}%
        \vspace{-1em}

  \label{sotacomparison}%
\end{table}%

\subsection{Experimental Results}
\subsubsection{Results on Image Segmentation}
\textbf{Weakly supervised segmentation:} Table~\ref{sotacomparison} shows comparison of recently introduced weakly supervised semantic segmentation methods with various levels of supervision. These methods all use a segmentation model based on VGG-16~\cite{simonyan2014very}. Our method achieves mIoU values of 63.9 and 65.0 for PASCAL VOC 2012 validation and test images respectively, which is 94.4\% of that of DeepLab~\cite{chen2014semantic}, trained with fully annotated data, which achieved an mIoU of 67.6 on validation images.
Our method is 3.1\% better on test images than the best method which uses only image-level annotations for supervision.
Our method also significantly outperformed several methods which have additional, as well as image-level annotations. These methods include TransferNet~\cite{hong2016learning} which is trained on pixel-level annotations of 60 classes not included in the PASCAL VOC. AISI~\cite{hu2018associating} has a salient instance detector, which is trained on well-annotated instance-level saliency maps, which are one of the most difficult forms of annotation to obtain. 
Our method also thoroughly outperformed existing methods based on the ResNet backbone~\cite{he2016deep} as shown in Table~\ref{resnets}. 

Figure~\ref{segsample} shows some examples of predicted segmentation masks produced by Bootstrap-Web~\cite{shen2018bootstrapping}, FickleNet~\cite{lee2019ficklenet}, and our system, using both VGG-16 and ResNet-based segmentation models. 
In general, our proxy ground truth covers larger regions of the target object than those produced by other methods, so that the segmentation masks produced by our method tend to be more accurate.
\begin{table}[tbp]
  \centering
  \caption{Comparison of the performance of weakly supervised segmentation
methods using the ResNet-based segmentation model on the VOC 2012 validation and test sets. \\[-2em]}
    \begin{adjustbox}{max width=0.48\textwidth}
    \begin{threeparttable}

    \begin{tabular}{lccc}
    \Xhline{1pt}
        \\[-0.9em]
    Methods & Backbone & ~\textit{val} ~  & ~\textit{test} ~\\
    \hline\hline    \\[-0.9em]
    \multicolumn{4}{l}{Weakly supervised methods:}\\
    MCOF~\cite{wang2018weakly}& ~~ResNet 101 ~ & ~60.3~ & 61.2~\\
    DCSP~\cite{chaudhry2017discovering}& ResNet 101~&  ~ 60.8  ~  &  ~61.9 ~\\
    DSRG~\cite{huang2018weakly}& ResNet 101~&  ~ 61.4  ~  &  ~63.2 ~\\
    AffinityNet~\cite{ahn2018learning}& ResNet 38~ &  ~ 61.7  ~  & ~63.7 ~\\
    SeeNet~\cite{hou2018self} &~ResNet 101~ & ~  63.1  ~  &62.8  ~\\
    CIAN~\cite{fan2018cian} &~ResNet 101~ & ~  64.1  ~  &64.7  ~\\

FickleNet~\cite{lee2019ficklenet} &~ResNet 101~ & ~  64.9  ~  &65.3  ~\\
    \hline\\[-0.9em]
    \multicolumn{4}{l}{Webly supervised methods:}\\
    Boot-Web~\cite{shen2018bootstrapping} &~ResNet 50~ & ~  63.0  ~  &63.9  ~\\
    Ours &~ResNet 101~ & ~  \textbf{66.5}  ~  &~ \textbf{67.4}  ~\\
    \Xhline{1pt}
    \end{tabular}%
    \end{threeparttable}
    \end{adjustbox}
  \label{resnets}%
  \vspace{-0.3em}
\end{table}%

\begin{table}[tbp]
  \centering
  \caption{Comparison of webly supervised segmentation methods on VOC 2012 validation and test images. The `Samples' column contains the total number of samples used for training, including the VOC images. \\[-2em]}
    \begin{adjustbox}{max width=0.48\textwidth}

    \begin{tabular}{lccc}
    \Xhline{1pt}\\[-0.95em]
    Methods   & Samples & val   & test \\
    \hline\hline
    \\[-0.9em]
    \multicolumn{4}{l}{Methods using web images:}\\
     $\text{WebS-i2}_{\text{~~CVPR '17}}$~\cite{jin2017webly}      & 20.3K  & 53.4  & 55.3 \\

    $\text{Boot-Web}_{\text{~~CVPR '18}}$~\cite{shen2018bootstrapping}      & 87.3K & 58.8  & 60.2 \\
    \hline\\[-0.9em]
    \multicolumn{4}{l}{Methods using web videos:}\\
         $\text{M-CNN}_{\text{~~ECCV '16}}$~\cite{tokmakov2016weakly}      & 13.6K  & 38.1  & 39.8 \\

    $\text{Web-Crawl}_{\text{~~CVPR '17}}$~\cite{hong2017weakly}      & 971K  & 58.1  & 58.7 \\
   Ours      & 25.5K & \textbf{63.9}  &  \textbf{65.0}\\
    \Xhline{1pt}
    \end{tabular}%
    \end{adjustbox}
  \label{sotaweb}%
  \vspace{-1em}

\end{table}%

\begin{figure*}[t]
  \centering
  \includegraphics[width=0.97\linewidth]{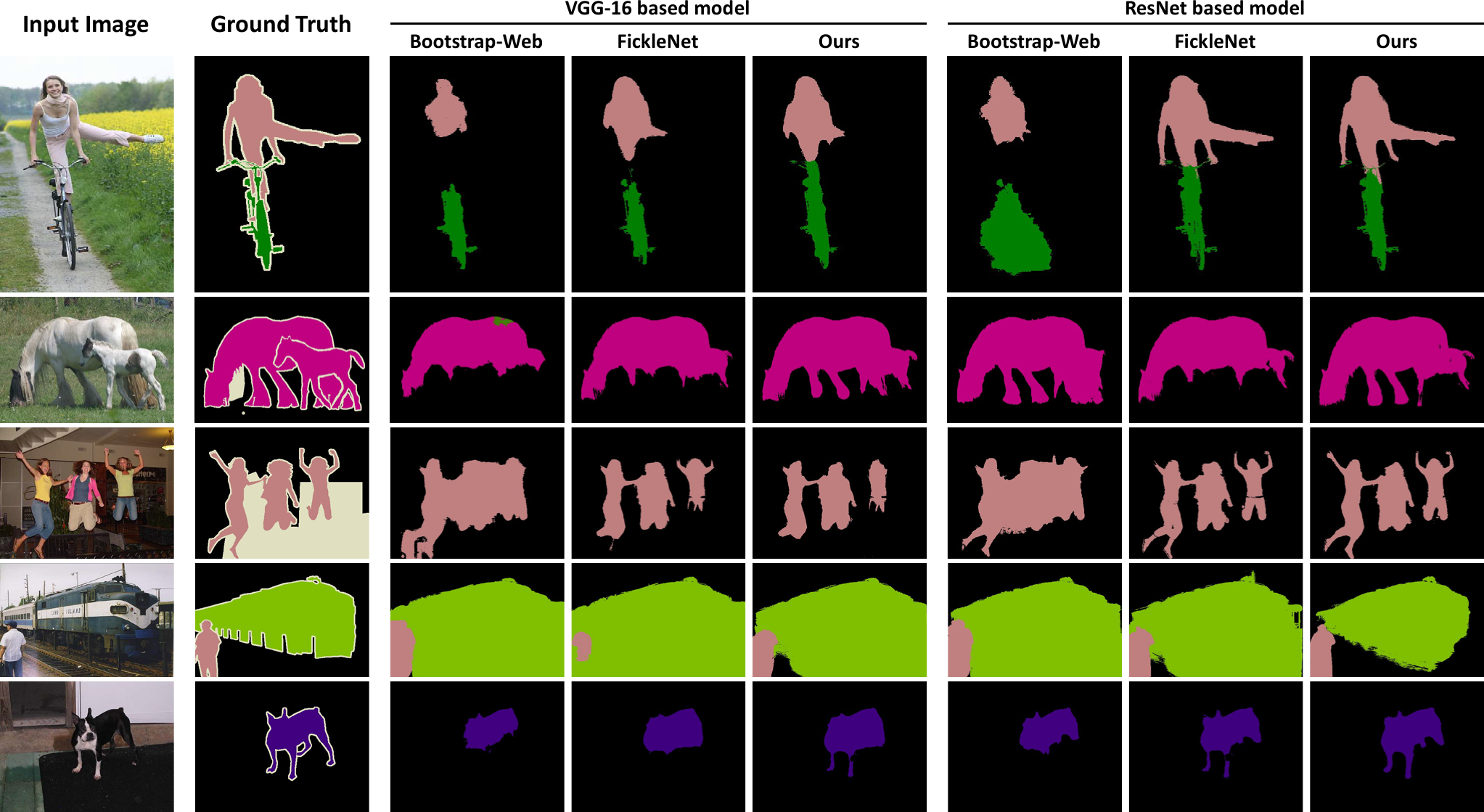}\\[-0.7em]
  \caption{\label{segsample} Examples of predicted segmentation masks for PASCAL VOC 2012 validation images.}\vspace{-0.8em}
\end{figure*}

\begin{figure*}
\begin{minipage}{0.4\linewidth}

\centering
  \captionof{table}{Comparison of video object segmentation methods with various supervision on the YouTube-Object dataset. \\[-2em]}
      \begin{adjustbox}{max width=\textwidth}
  \begin{threeparttable}

    \begin{tabular}{lcccc}
        \Xhline{1pt}

    Method &  sup. & mIoU  \\
    \hline\hline
    $\text{Tang }\textit{et al.}_{\text{~~CVPR '13}}$~\cite{tang2013discriminative}     &   $\mathcal{U}$& 23.9  \\
    $\text{Papazoglou }\textit{et al.}_{\text{~~ICCV '13}}$~\cite{papazoglou2013fast}     &  $\mathcal{U}$ & 46.8   \\
     $\text{Jang }\textit{et al.}_{\text{~~CVPR '16}}$~\cite{jang2016primary}    &  $\mathcal{U}$ & 53.0     \\
    $\text{Zhang }\textit{et al.}_{\text{~~CVPR '15}}$~\cite{zhang2015semantic}&$\mathcal{B}$  & 54.1   \\
     $\text{Drayer }\textit{et al.}_{\text{~~ArXiv '16}}$~\cite{drayer2016object}    &$\mathcal{B}$   & 56.2   \\
        $\text{Zhang }\textit{et al.}_{\text{~~TPAMI '18}}$~\cite{zhang2018semantic}  &$\mathcal{B}$  & 61.7   \\
        $\text{Saleh }\textit{et al.}_{\text{~~ICCV '17}}$~ \cite{saleh2017bringing}  &$\mathcal{I}$ & 53.3  \\

     $\text{Web-Crawl}_{\text{~~CVPR '17}}$~\cite{hong2017weakly}  &$\mathcal{I}$  & 58.6   \\
     $\text{SROWN}_{\text{~~TIP '18}}$~\cite{yang2018segmentation}  &$\mathcal{I}$  &61.9   \\
    Ours  &$\mathcal{I}$ & \textbf{62.1} \\
        \Xhline{1pt}

    \end{tabular}%
    \begin{tablenotes}
  \footnotesize
            \item $\mathcal{U}-$Unsupervised, $\mathcal{B}-$Bounding boxes, $\mathcal{I}-$Image labels
        \end{tablenotes}
     \end{threeparttable}
    \end{adjustbox}
  \label{table_videoseg}%
  
\end{minipage}\hfill
\begin{minipage}{0.58\linewidth}
\centering
  \includegraphics[width=\linewidth]{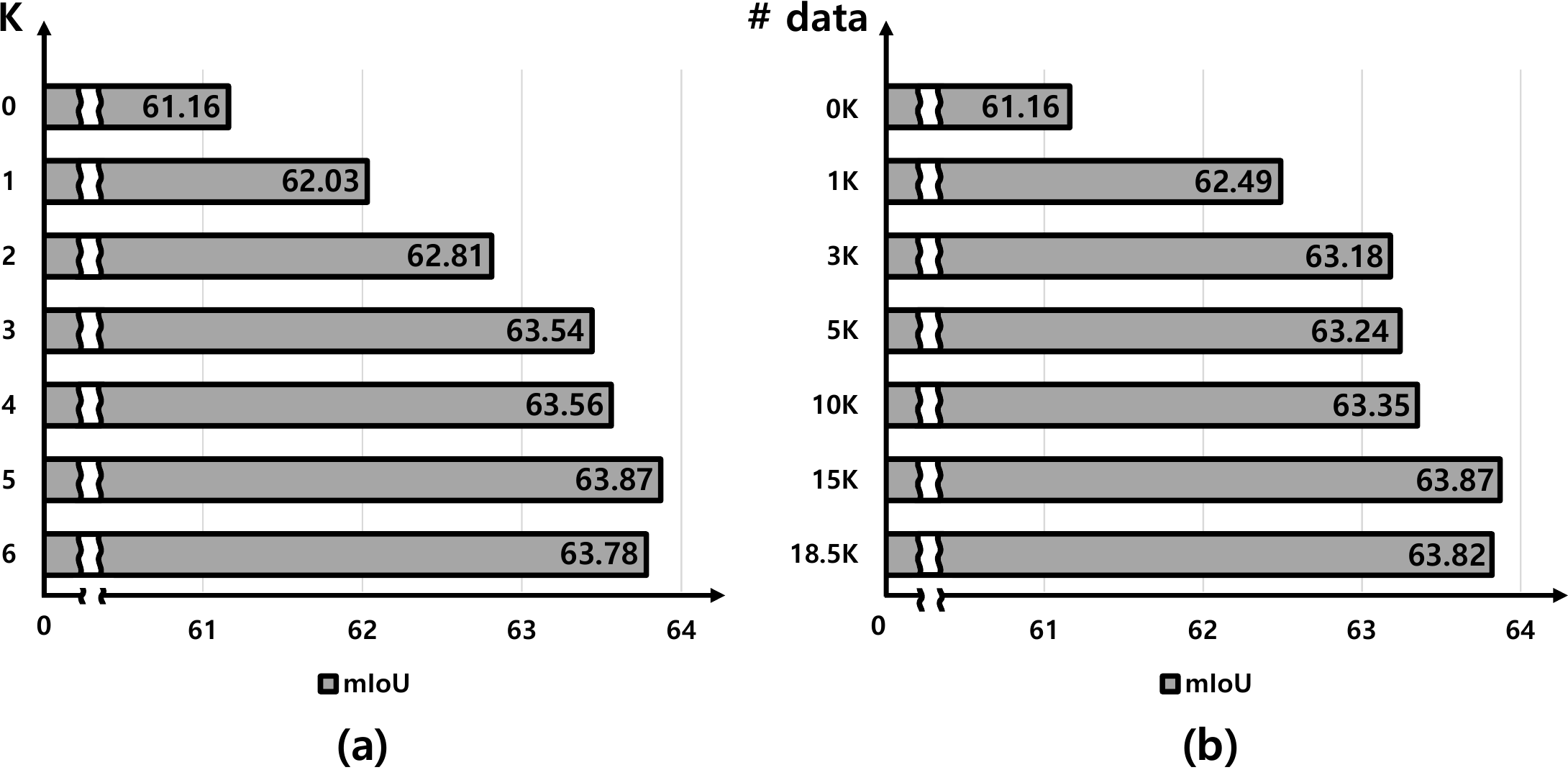} \\[-0.7em]
  \captionof{figure}{(a) Comparison of mIoU scores using different $K$. $K=0$ denotes the result trained without web videos. (b) Comparison of mIoU scores using different numbers of web data.}
\label{ablation}

\end{minipage}
\\[-0.97em]
\end{figure*}

\noindent\textbf{Webly supervised segmentation:} 
Table~\ref{sotaweb} shows mIoU values achieved on the PASCAL VOC 2012 dataset of webly supervised segmentation methods and the total number of training samples of each method.
Our method showed the best performance despite being trained on a relatively small amount of data: 10.5k PASCAL VOC images and 15k frames of web video, whereas Web-Crawl~\cite{hong2017weakly} and Boot-Web~\cite{shen2018bootstrapping} used 971k and 87.3k samples for training, respectively.\\[-1.5em]
\vspace{-0.3em}
\subsubsection{Results on Video Segmentation}
\begin{figure*}[t]
  \centering
  \includegraphics[width=0.95\linewidth]{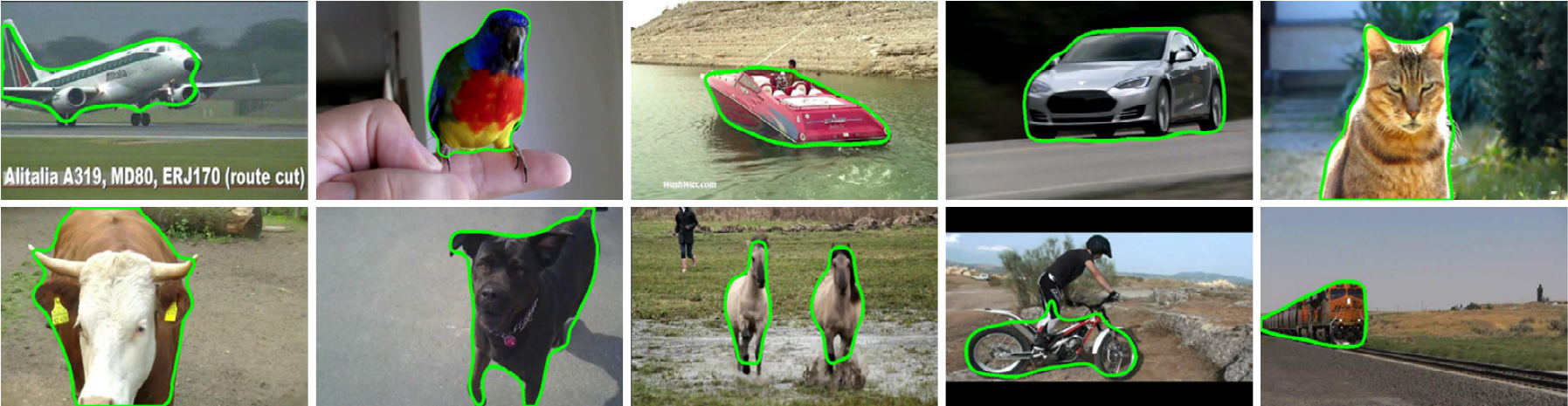}\\[-0.7em]
  \caption{\label{videoseg_samples} Predicted masks for frames of the YouTube-Object dataset. Segmented regions are outlined by green curves.}\vspace{-0.8em}
\end{figure*}

In Table~\ref{table_videoseg}, we assessed the segmentation results produced by our system on the YouTube-Object dataset~\cite{prest2012learning}, and compared with state-of-the-art video segmentation methods with various degrees of supervision.
We used segmentation masks annotated by Jain \textit{et al.}~\cite{jain2014supervoxel} for ground-truth of evaluation. 
We also report the mIoU of DSRG~\cite{huang2018weakly} and FickleNet~\cite{lee2019ficklenet} trained on $\mathcal{I}$ alone as baselines. Our method showed better performance than the existing methods and even surpassed the methods which use stronger supervision, such as bounding boxes.
A few examples of predicted segmentation masks for the YouTube-Object dataset are shown in Figure~\ref{videoseg_samples}.

\begin{figure*}[t]
  \centering
  \includegraphics[width=0.99\linewidth]{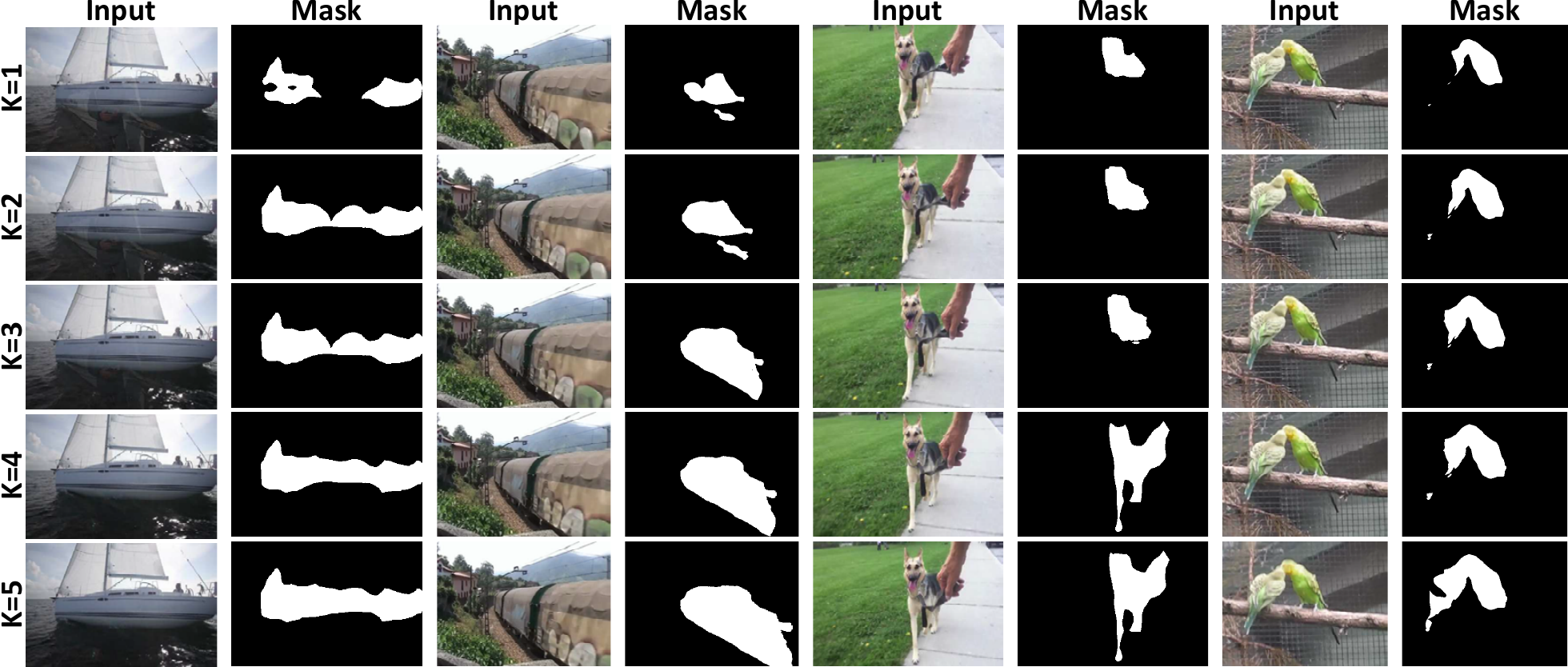}\vspace{-0.5em}
  \caption{\label{different_K} Examples of aggregated mask at each incremental warping step.\\[-1em]}
\end{figure*}
\subsection{Ablation Study}


\noindent\textbf{Number of frames aggregated:}
Figure~\ref{ablation}(a) shows mIoU scores for the PASCAL VOC 2012 validation images with different number of aggregated frames $K$.
Using a single frame ($K=1$) results in only slight performance improvement on the score without any web videos ($K=0$).
Increasing the number of successive frames across which maps are aggregated improves performance, which we expect, as larger regions of the target are represented by the final mask. But we find that aggregation above $K=5$ is not beneficial. This can be attributed to the approximations of the optical flow involved in frame-to-frame warping. Errors can be expected to build up as regions are warped across more frames, undermining the accuracy of the proxy ground truth (see Appendix).
The examples of aggregated mask at each incremental warping step are shown in Figure~\ref{different_K}.
 

\noindent\textbf{Number of web samples:}
The effect of the number of web samples is shown in Figure~\ref{ablation}(b).
Without any web videos, FickleNet~\cite{lee2019ficklenet} is trained with the PASCAL VOC data alone, and the mIoU is 61.2.
The mIoU value increases monotonically up to 15,000 samples. More samples produce little change in the performance.

\noindent\textbf{Other weakly supervised segmentation networks:} 
In addition to FickleNet~\cite{lee2019ficklenet}, we experimented with SEC~\cite{kolesnikov2016seed} and DSRG~\cite{huang2018weakly} with our method.
Table~\ref{table_segnet} shows the performance of those three segmentation networks with image data $\mathcal{I}$ alone, with an additional video dataset $\mathcal{V}$, and also with domain adaptation $\mathcal{DA}$.
SEC~\cite{kolesnikov2016seed} does not have a retraining process, so we add a retraining step, before domain adaptation, along the line of that in DSRG~\cite{huang2018weakly}.
The results in Table~\ref{table_segnet} show that our method works effectively with the three weakly supervised semantic segmentation networks.

\begin{table}[tbp]
\setlength{\tabcolsep}{3pt}

  \centering
  \caption{Effects of addingvideo data $\mathcal{V}$ and domain adaptation $\mathcal{DA}$ on three weakly supervised segmentation models. \\[-2em]
}
    \begin{adjustbox}{max width=0.48\textwidth}
  \begin{threeparttable}

    \begin{tabular}{rcccc}
    \Xhline{1pt}\\[-0.95em]
          &       & SEC~\cite{kolesnikov2016seed} &  DSRG~\cite{zhang2018semantic} & FickleNet~\cite{lee2019ficklenet} \\
    \hline\hline\\[-0.95em]
    \multicolumn{1}{c}{\multirow{3}[0]{*}{mIoU}} & $\mathcal{I}$ &    50.7   &    59.0   &  61.2\\
    \multicolumn{1}{c}{} & $\mathcal{I}$+$\mathcal{V}$ &   ~ 59.5$^1$  & 62.1      & 63.2 \\
    \multicolumn{1}{c}{} & $\mathcal{I}$+$\mathcal{V}$+$\mathcal{DA}$ &   61.1   &  62.9    & 63.9 \\
    \Xhline{1pt}
    \end{tabular}%
    \begin{tablenotes}
  \footnotesize
            \item $^1$ A retrain process is included\\[-3em]
        \end{tablenotes}
     \end{threeparttable}
    \end{adjustbox}
  \label{table_segnet}%
\end{table}%

The results for SEC~\cite{kolesnikov2016seed} offer the possibility of a fairer comarison with Bootstrap-Web~\cite{shen2018bootstrapping}, which is based on SEC.
For the PASCAL VOC validation images, our method achieved mIoU values of 61.1, while Bootstrap-Web~\cite{shen2018bootstrapping} achieved 58.8. 

\vspace{-0.5em}
\section{Conclusions}
We have proposed a method to use videos automatically obtained from the web as additional data in weakly supervised semantic segmentation. 
We obtain activated regions from each frame of a video and aggregate them on a single image, so that our proxy ground truth covers large regions of the target object.
This method does not require additional supervision, it can be realized without complicated optimization processes or off-the-shelf segmentation methods, and it requires relative few samples, because a lot of information extracted from many frames can be aggregated into a single frame.
We have demonstrated that our method produces better results than those from other state-of-the-art weakly and webly supervised approaches. We have also demonstrated that our method works effectively with several weakly supervised semantic segmentation networks.


\bigskip
\noindent\textbf{Acknowledgements:}
This work was supported by the National Research Foundation of Korea (NRF) grant funded by the Korea government (MSIT) [2018R1A2B3001628], AIR Lab (AI Research Lab) in Hyundai Motor Company through HMC-SNU AI Consortium Fund, Institute for Information \& communications Technology Planning \& Evaluation (IITP) grant funded by the Korea government (MSIT) (No.2019-0-01367
), Samsung Electronics (DS and Foundry), and the Brain Korea 21 Plus Project in 2019.

\setcounter{section}{0}
\renewcommand\thesection{\Alph{section}}
\setcounter{table}{0}
\renewcommand{\thetable}{A\arabic{table}}
\setcounter{figure}{0}
\renewcommand{\thefigure}{A\arabic{figure}}

\onecolumn
\section{Appendix}
\vspace{-1em}
\begin{figure}[h]
\begin{minipage}{0.5\textwidth}
\noindent\textbf{Effects of Conditional Random Field (CRF):}
Table~\ref{without_crf} shows mIoU values on PASCAL VOC 2012 validation images with and without CRF. 

\noindent\textbf{Per-class Results:} Table~\ref{class-specific-results} shows the per-class mIoU of our method and baselines. 

\noindent\textbf{Qualitative Results:}
Figure~\ref{segsample} shows additional examples of predicted segmentation mask produced by DSRG~\cite{huang2018weakly}, Bootstrap-Web~\cite{shen2018bootstrapping}, FickleNet~\cite{lee2019ficklenet}, and our system, using both VGG-16 and ResNet-based segmentation models.

The examples of aggregated mask at each incremental warping step are shown in Figure~\ref{warpsample}. We present a failure case in the last row. 

\end{minipage}\hfill
\begin{minipage}{0.46\textwidth}
  \centering
  \captionof{table}{Quantitative results with and without CRF. Diff. denotes the difference with and without CRF.}\vspace{-1em}
    \begin{tabular}{lccc}
    \Xhline{1pt}\\[-0.95em]

    \multicolumn{1}{c}{\multirow{2}[0]{*}{Method}} & \multicolumn{2}{c}{mIoU} & \multicolumn{1}{c}{\multirow{2}[0]{*}{Diff.}} \\
    
    \multicolumn{1}{c}{} & ~with CRF~ & without CRF ~& \multicolumn{1}{c}{} \\
    \hline\hline\\[-0.95em]
    Ours & \textbf{63.9} & \textbf{62.8} & \textbf{1.1} \\
    SeeNet~\cite{hou2018self} & 61.1 & 59.9 & 1.2\\
    FickleNet~\cite{lee2019ficklenet} &  61.2     &  59.7     &  1.5\\
    DSRG~\cite{huang2018weakly}   &    59.0  &   56.5     &   2.5 \\
    MDC~\cite{wei2018revisiting}  &  60.4     &    57.1   &   3.3 \\
    GAIN~\cite{li2018tell}  &   55.3    &  50.8     &   4.5 \\
    \Xhline{1pt}
    \end{tabular}%
  \label{without_crf}%

\end{minipage}
\vspace{-1em}
\end{figure}

\begin{figure*}[h]
  \centering
  \vspace{-1em}
  \includegraphics[width=0.95\linewidth]{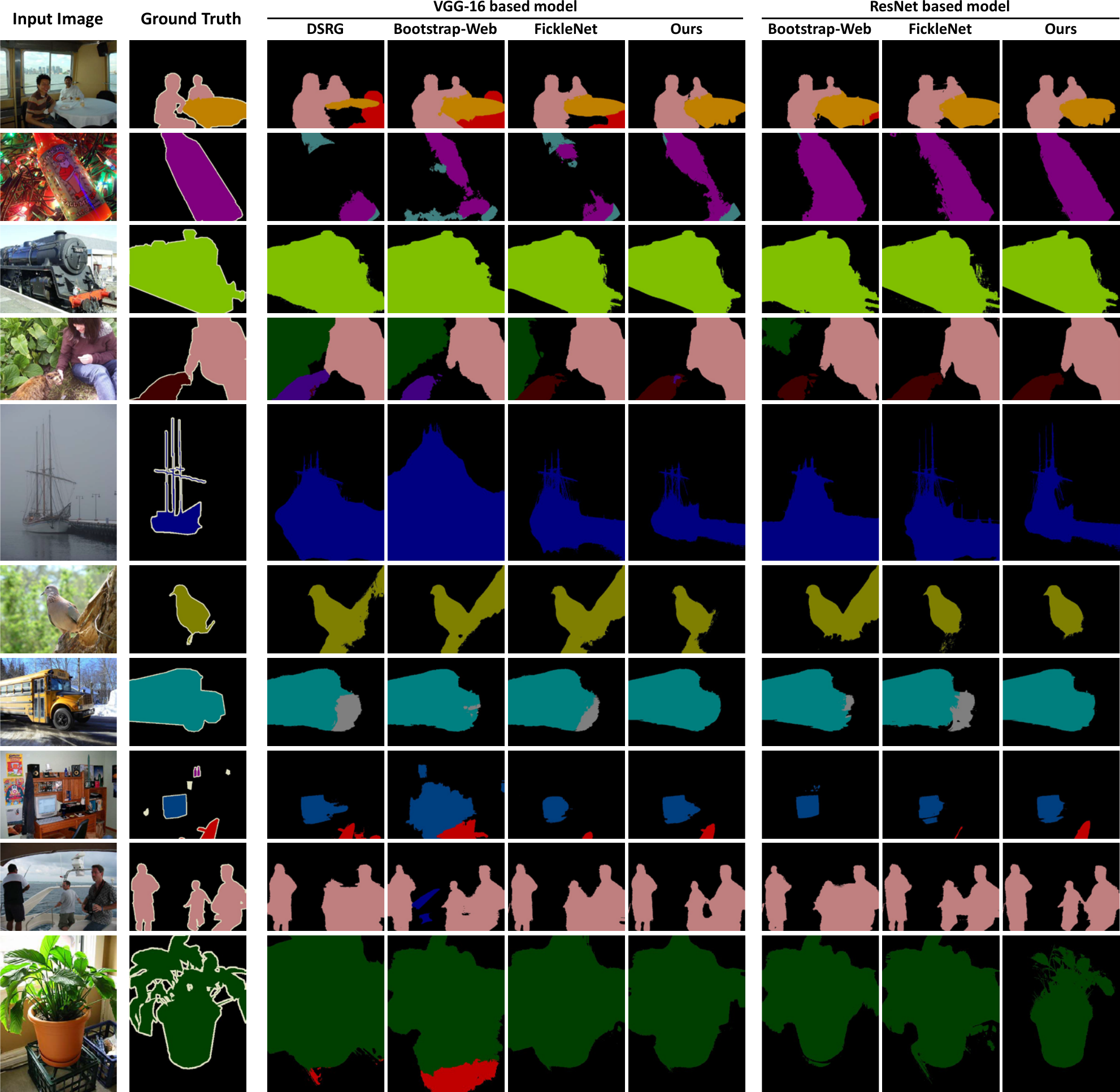}
  \vspace{-0.9em}
  \caption{\label{segsample} Examples of predicted segmentation mask for PASCAL VOC 2012 validation images.}
\end{figure*}

\twocolumn
\renewcommand{\tabcolsep}{2pt}

\begin{table*}[!ht]
  \caption{Comparison of per-class mIoU scores.}
    \vspace{-0.8em}
  \centering
  \begin{adjustbox}{max width=\textwidth}
    \begin{tabular}{lccccccccccccccccccccc|c}
    
        \Xhline{1pt}

        \\[-0.95em]
    Rates~~~ & bkg& aero  & bike  & bird  & boat  & bottle & bus   & car   & cat   & chair & cow   & table & dog   & horse & motor & person & plant & sheep & sofa  & train & tv  ~ & mIOU \\
   
    \hline
    \hline
    \\[-0.9em]
   \multicolumn{22}{l}{Results on validation images:}\\
SEC &   82.4    &   62.9    &   26.4    &   61.6    &   27.6    &    38.1   &    66.6   &   62.7    &   75.2    &    22.1   &   53.5    &   28.3    &    65.8  &    57.8   &   62.3    &    52.5   &   32.5    &   62.6    &   32.1    &    45.4   &    45.3  ~ &  50.7\\

Ours (SEC) &   88.0    &   80.1    &   32.7    &   71.8    &   46.9    &    60.9   &    79.0   &   73.5    &   78.8    &    25.6   &   66.8    &   42.1    &    74.4   &    63.6   &   70.0    &    68.2   &   42.1    &   69.3    &   36.9    &    62.3   &    50.8  ~ &  61.1\\

    DSRG &   87.5    &   73.1    &   28.4    &   75.4    &   39.5    &    54.5  &   78.2    &   71.3    &   80.6    &   25.0    &   63.3    &    25.4   &   77.8    &    65.4   &   65.2    &   72.8    &   41.2    &    74.3   &   34.1    &   52.1    &   53.0   ~ &  59.0\\
    
 Ours (DSRG) &   89.7    &   81.8    &  33.2    &   75.7    &   61.4    &    60.8  &   80.3    &   74.9    &   80.0    &   24.6    &   70.0   &    35.1   &   75.4    &    68.3   &   66.6    &   71.6    &   42.4    &    74.7   &   36.8    &   64.4    &   52.6   ~ &  62.9\\
 
    FickleNet &   88.1    &   75.0    &   31.3    &   75.7    &   48.8   &   60.1    &    80.0  &    72.7   &    79.6   &  25.7     &   67.3    &   42.2    &    77.1&    67.5   &   65.4    &  69.2 &  42.2   &    74.1  & 34.2  &  53.7 &   54.7~ &  61.2\\
    
     Ours (w/o. CRF) &   89.4    &   80.0    &   34.5    &   74.5    &   60.7   &   61.1    &    79.3  &    73.3   &    77.2   &  21.3     &   68.3    &   44.4    &    74.2&    67.2   &   68.6    &  69.3 &  43.2   &    71.3  & 39.0  &  68.3 &   51.7~ &  62.8\\
     Ours (VGG-16) &   89.8    &   81.8    &   35.5    &   75.4    &   63.5   &   61.2    &    80.0  &    73.1   &    79.2   &  21.7     &   70.9    &   46.8    &    76.7&    70.1   &   69.3    &  69.5 &  41.8   &    74.2  & 39.5  &  68.3 &   52.8~ &  63.9\\
    Ours (ResNet 101) &   90.8    &   82.2    &   35.1    &  82.4    &   72.2   &   71.4    &    82.7  &    75.0   &    86.9   &  18.3     &   74.2    &   29.6    &    81.1&    79.2   &   74.7    &  76.4 &  44.2   &    78.6  & 35.4  &  72.8 &   63.0~ &  66.5\\
    \hline
    \multicolumn{22}{l}{Results on test images:}\\
    Ours (VGG-16) &   90.3    &   77.0    &   35.2    &   76.0    &   54.2   &   64.3    &    76.6  &    76.1   &    80.2   &  25.7     &   68.6    &   50.2    &    74.6&    71.8   &   78.3    &  69.5 &  53.8   &    76.5  & 41.8  &  70.0 &   54.2~ &  65.0\\
    Ours (ResNet 101) &   91.2    &   84.2    &   37.9    &   81.6    &   53.8   &   70.6    &    79.2  &    75.6   &    82.3   &  29.3     &   76.2    &   35.6    &    81.4&    80.5   &   79.9    &  76.8 &  44.7   &    83.0  & 36.1  &  74.1 &   60.3 ~ &  67.4\\
        \Xhline{1pt}

    \end{tabular}%
  \end{adjustbox}%
  \label{class-specific-results}%
\end{table*}%

\begin{figure*}[t]
  \centering
  \includegraphics[width=0.92\linewidth]{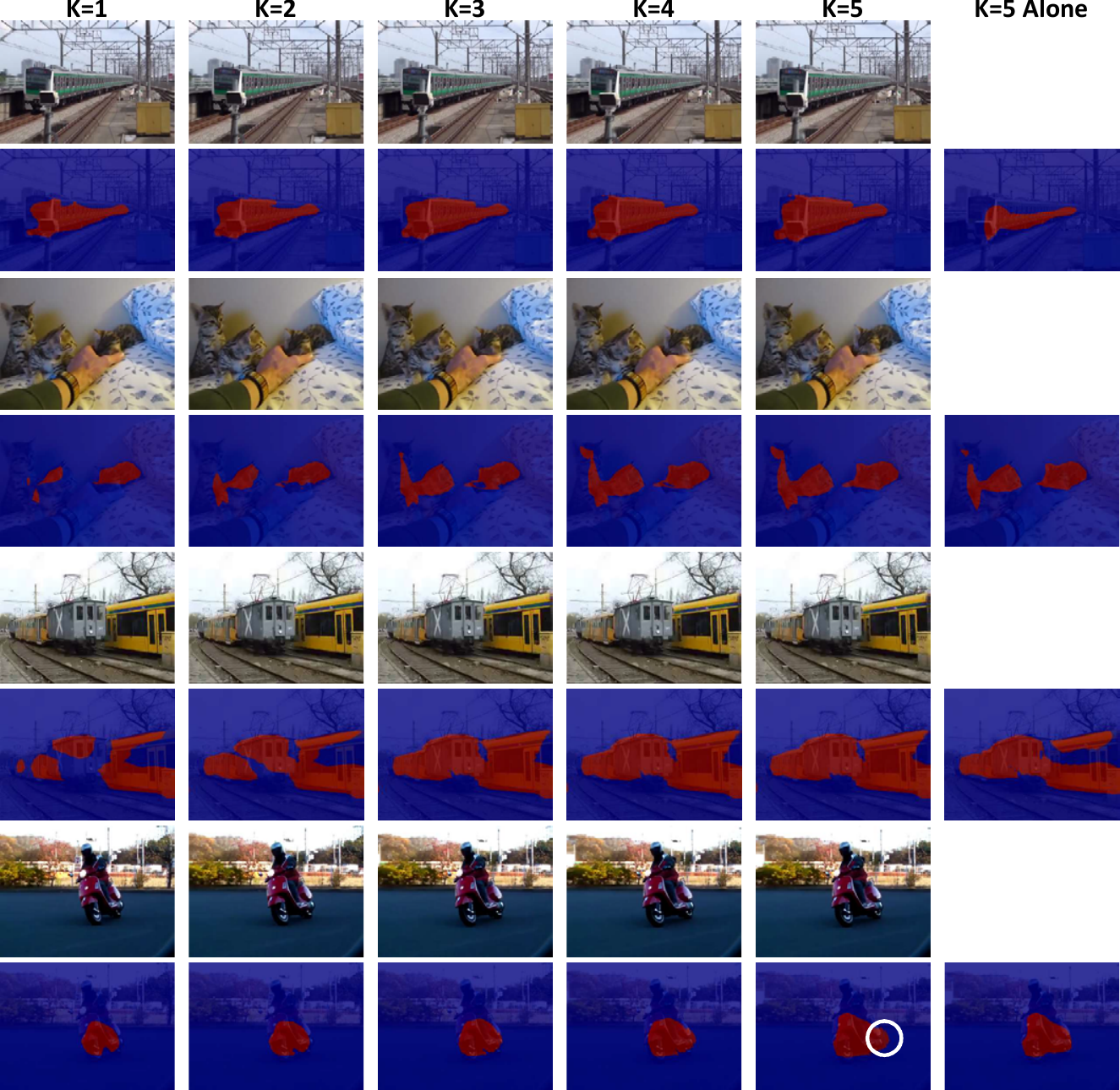}
  \vspace{-0.8em}
  \caption{\label{warpsample} Examples of aggregated mask at each incremental warping step, and ``K=5 Alone" denotes the mask from the last frame. The white circle in the last row indicates an error resulting from incremental warping.}
\end{figure*}

\clearpage

{\small
\bibliographystyle{ieee}
\bibliography{egbib}
}

\end{document}